\RequirePackage{amsmath}
\documentclass[runningheads]{llncs}

\usepackage[T1]{fontenc}
\usepackage{CJK}
\usepackage{graphicx}
\usepackage{tabularx}
\usepackage{mdframed}
\usepackage{xcolor}

\usepackage{balance}
\usepackage{multicol}
\usepackage{amssymb}

\usepackage{hyperref}
\hypersetup{
    colorlinks=true,
    linkcolor=blue,
    citecolor=blue,
    filecolor=magenta,      
    urlcolor=blue
    }
\usepackage{pifont}
\newcommand{\cmark}{\ding{51}}%
\newcommand{\xmark}{\ding{55}}%
\usepackage[ruled,vlined]{algorithm2e}

\usepackage{inconsolata}
\usepackage[
  subtle
]{savetrees}
\usepackage{enumitem}
\newcommand{\BI}{BI }
\newcommand{\BIN}{BI}
\usepackage{listings}
\lstdefinelanguage{json}{
    basicstyle=\normalfont\ttfamily,
    showstringspaces=false,
    breaklines=true,
    frame=single
}
\newcommand{\mintsql}[1]{\lstinline{#1}}
\newcommand{\sqlbreak}{\\ \hspace*{10ex}}
\newcommand\blfootnote[1]{%
  \begingroup
  \renewcommand\thefootnote{}\footnote{#1}%
  \addtocounter{footnote}{-1}%
  \endgroup
}

\begin{document}

\title{BIS: NL2SQL Service Evaluation Benchmark for Business Intelligence Scenarios}
\authorrunning{Caglayan et al.}
\author{Bora Caglayan\inst{1}, Mingxue Wang\inst{1}, John D. Kelleher\inst{2}, Shen Fei\inst{3} \\
Gui Tong\inst{3}, Jiandong Ding\inst{3}, Puchao Zhang\inst{1}  }
\institute{
Huawei Ireland Research Centre, Dublin, Ireland \\
\email{\{bora.caglayan, wangmingxue, zhangpuchao\}@huawei.com} 
\and
ADAPT Research Centre, \\ School of Computer Science and Statistics,
Trinity College Dublin, Dublin, Ireland \\
\email{john.kelleher@tcd.ie} 
\and
Huawei Technologies Co., Ltd.\\
\email{\{shenfei8, guitong, dingjiandong2\}@huawei.com} 
}


\maketitle

\begin{abstract}
NL2SQL (Natural Language to Structured Query Language) transformation has seen wide adoption in Business Intelligence (\BIN) applications in recent years. However, existing NL2SQL benchmarks are not suitable for production \BI scenarios, as they are not designed for common business intelligence questions. To address this gap, we have developed a new benchmark focused on typical NL questions in industrial \BI scenarios. We discuss the challenges of constructing a \BI-focused benchmark and the shortcomings of existing benchmarks. Additionally, we introduce question categories in our benchmark that reflect common \BI inquiries. Lastly, we propose two novel semantic similarity evaluation metrics for assessing NL2SQL capabilities in \BI applications and services.
\end{abstract}

\section{Introduction}

NL2SQL enables users to ask questions using natural language with no or little knowledge of SQL query composition or database schema details \cite{date1989guide}, \cite{kim2020natural}, \cite{katsogiannis2023survey}. It provides analysis flexibility for non technical business analysts and has started to be adapted as a feature for BI applications. Over the years, multiple benchmarks have been proposed to evaluate NL2SQL models \cite{yu-etal-2018-spider}, \cite{zhongSeq2SQL2017}. However, the question types, data schemas and sample database contents of these benchmarks are not designed for common \BI scenarios. For example, the  WikiSql \cite{zhongSeq2SQL2017} benchmark mostly contains factual questions over Wikipedia and each query is executed on a single table. In addition, widely used performance measures in NL2SQL evaluate only the match rate of queries as either a perfect match or a no match. These performance measures overly penalize partial matches that may still contain valuable information for the \BI analysts. 
\blfootnote{\scriptsize This work is partly funded by the ADAPT Research Centre for AI-Driven Digital Content Technology, which is funded by Research Ireland through the Research Ireland Centres Programme and is co-funded under the European Regional Development Fund (ERDF) through Grant 13/RC/2106 P2.}

NL2SQL can be considered as a use case for no-code software development.
No-code is an approach in software development aimed at removing the need for manual coding   \cite{rokis2022challenges} \cite{elbatanony2021towards}. 
The no-code approach seeks to improve efficiency in organizations by replacing manual coding requirements during various stages of software development with features that do not require coding. 
This approach also provides secondary efficiency benefits by enabling domain experts to apply their knowledge without relying on developers.
NL2SQL can be regarded as a no-code service for interfacing with databases, potentially replacing repetitive manual dashboard-building tasks and thereby empowering end users while improving efficiency. By transitioning the database interface from SQL to natural language, NL2SQL enables the creation of a pipeline for other data services using its output, such as automated dashboards, chatbots, and data visualization tools.

In this paper, we propose a new benchmark named BIS to evaluate NL2SQL models focusing on common questions and database schemas observed in \BI scenarios. 
Our contributions can be summarized as follows: \textbf{1)} We describe the shortcomings of the existing benchmarks and evaluation metrics for NL2SQL models in BI scenarios; \textbf{2)} We propose a new benchmark with two novel evaluation metrics (semantic query similarity and result similarity) to address the shortcomings.  

The rest of the paper is organized as follows: In Section \ref{section:challenge}, we discuss the common challenges of existing NL2SQL benchmarks for business intelligence query scenarios. In Section \ref{section:questions}, we outline the major categories of business intelligence queries commonly used in our organization. In Section \ref{section:benchsetup}, we provide a description of the dataset, including its download instructions and usage guidelines. In Section \ref{section:metric}, we propose two novel accuracy metrics that assess the partial similarity of queries to evaluate model performance more realistically. Finally, we conclude the paper with a discussion on ethics and data protection, as well as the limitations of the current benchmark.

\section{NL2SQL Benchmarks and Their Challenges for BI Applications}
\label{section:challenge}

\begin{table}[t] \scriptsize
\caption{Open Source Benchmark Comparison for NL2SQL Benchmarks}
\label{table:open-benchmarks}
\begin{tabularx}{\textwidth}{ |X|X|X|X|X|r| } 
 \hline
\textbf{Benchmark} & \textbf{Definition}  & \textbf{BI Question Categories} & \textbf{SQL Query Evaluation Metrics} & \textbf{SQL Result Evaluation Metrics}& \textbf{Question \#}\\
 \hline
 WikiSql \cite{zhongSeq2SQL2017} & Generated questions from Wikipedia  &  \xmark  & SQL logical form exact match & Result exact match & 80645 \\ 
\hline
 Spider \cite{yu-etal-2018-spider} & Student generated database covering a wide range of domains   & \xmark&  SQL component exact match & Result exact match &  10181\\ 
 \hline
 Yelp \& IMDB \cite{data-sql-imdb-yelp} & Yelp website and the Internet Movie Database   & \xmark &  \xmark&  \xmark & 259 \\ 
 \hline
MAS \cite{mas} & Microsoft academic search database   & \xmark &  \xmark &  \xmark & 196 \\ 
 \hline
Advising \cite{advising} & Course information & \xmark& SQL component exact match & \xmark & 4579\\
 \hline
CSpider \cite{cspider} & Spider Benchmark for Chinese   & \xmark &   SQL component exact match & \xmark & 9691\\
 \hline
 BIS (Our benchmark) &  Benchmark with temporal information for BI applications    & \cmark &  SQL semantic similarity & Result partial similarity & 239\\
 \hline
\end{tabularx}
\end{table} 

Over the years many NL2SQL benchmarks have been proposed. A non-comprehensive list of open NL2SQL benchmarks is provided in Table \ref{table:open-benchmarks}.
We refer readers to a recent survey by Qin et al. for a more comprehensive overview of the NL2SQL literature \cite{qin2022survey}.
Earlier benchmarks such as WikiSql contains queries on a single table \cite{zhongSeq2SQL2017}.
Recent benchmarks such as Spider contains  complex queries that operate on databases with relational structures \cite{cspider} \cite{yu-etal-2018-spider}.
In general, current benchmarks enable the evaluation of NL2SQL models in terms of exact and execution accuracy. 
However, some common NL2SQL challenges regarding \BI scenarios are not addressed or focused by the current benchmarks. 
We discuss these challenges in four groups below:

\subsection*{Challenges regarding database schemas} \BI databases may contain various irregularities in their schema definitions.
One irregularity is caused by identical data having different column names on different tables.
Such inconsistencies may either be caused by a lack of common naming guidelines across the organization or different naming conventions by different teams.
For example, \mintsql{task} and \mintsql{task_id} may contain the same information in two different tables. 
This is a challenge  for NL2SQL tasks, especially tasks that require joining multiple tables since most models only build a single semantic map with a column name.  
Therefore, mapping between different synonyms of the same column is essential in identifying the join operations for the tables. 
The other irregularity in schema design is caused by having identical column names on different tables for different data.
Developers may reuse the same schema templates to reuse the same names for different tables. 
For example,  \mintsql{metric_log_real} and \mintsql{metric_log_predicted} contain real and predicted values for a metric using the same column labels.
This is a  common design choice in operation monitoring systems as well, as usually the same set of metrics are collected from different sources
Existing benchmarks do not consider the effect of such schema irregularities on NL2SQL performance.

\subsection*{Challenges regarding database content} 
Previous NL2SQL benchmarks such as WikiSql\cite{zhongSeq2SQL2017} cover questions that query entities. They do not cover questions on time series data to the best of our knowledge \cite{qin2022survey}. However, in BI scenarios, multiple dimensional time series data and queries related to time series make up  most of the database. For example, \mintsql{metric_log_predicted} contains metrics with time interval associations. In \BI daily tasks, none or very few questions would be in the form of \textit{"What is the capital city of Ireland?"}. On the other hand, questions related to trends, trend comparisons and metric aggregations are quite common. Current benchmarks have no coverage of these common \BI questions. 

\BI database designers usually use metadata description tables to store information about data entities, such as error codes and filter keys. In these cases, user questions may not be clearly understood without linking the direct result with the metadata table even when the query itself does not  ask for the metadata table. For example, in our benchmark, \mintsql{o_rank_filter_vector} is described as the "vector engine ranking filter" in a metadata table. Other open benchmarks highlighted in Table \ref{table:open-benchmarks} do not cover these types of implicit metadata mapping scenarios to the best of our knowledge.

\subsection*{Challenges regarding question context} 
User questions of existing benchmarks do not cover complex temporal selections. However, in \BI queries most of the natural language questions have temporal constraints such as "on last Friday", "last 5 days", "compare with last week". Covering such time ranges in natural language questions are important for the evaluation.

More importantly, user questions are not categorized from the business analysis prospective such as questions related to the comparison of a metric between two groups or questions about metric trends. Hence, it is difficult for a detailed benchmark NL2SQL models to understand their performance on various categories of common \BI questions. More details of question categorization will be discussed in the Section \ref{section:questions}.

Technical terms are widely used in questions for \BI scenarios. As a consequence, domain knowledge is required to understand the user questions. For example, if a user asks about the percentage of people clicking on advertisements after seeing them, it requires some domain knowledge to understand that the user is asking about the the term "click through rate" (CTR) of a particular group of advertisements. As a result, the \mintsql{avg_ctr} column should be used to generate the SQL query.

\subsection*{Challenges regarding question language} 

\begin{CJK}{UTF8}{gbsn}
Issues in question context may also be related to usage of a combination of different languages in the same query. 
English abbreviation for technical terms are commonly used in combination with  Chinese or another non-Latin language.  For example, 平均 CVR 是多少 （i.e. what is average CVR). Non-English speaking users may use English and their native language together to query entities. In these scenarios, direct keyword matching between user questions and table information may not work.
\end{CJK}

\subsection*{Challenges regarding evaluation metrics} Performance measures proposed by current benchmarks usually employ the \textit{exact match} strategy to avoid evaluation complexities but this simplification might overlook partial or semantically identical predictions that might still be valuable for the user. Comparing two SQL statements is complicated due to two factors: 1) two syntactically almost completely different SQL statements might have same/similar meaning, 2) two different queries might have same/similar results.  We can illustrate the complexity of comparing predicted and actual queries with a simple example as follows:

\#1: \mintsql{SELECT count(*) FROM t GROUP BY day}
    
\#2: \mintsql{SELECT count(*) AS count FROM t GROUP BY day}

If we compare query \#1 and query \#2 using string distance based measure or exact query match, the accuracy would not be perfect.  However, from an execution point of view the queries can be considered the same. Such syntactic differences makes automated comparison even harder for more complex queries with join operations or complex aggregations. Syntax or edit distance based similarity measures commonly used to detect string similarity is not applicable due to several reasons. First, the same query can be expressed in different orders. Second, changes in queries do not have the same semantic effect. For example, some changes in queries such as different table selection is much more important than alias changes. For this reason, we defined two metrics based on SQL statement semantic similarity and SQL result accuracy as described in detail in Section \ref{section:metric}.

\subsection*{Challenges regarding predicted query performance}

For complex queries, the same result can be generated by many different queries with different performances. 
In addition for comparison or trend queries, queries can either be executed in a single step or as a combination of multiple queries.
Performance implications of such query combinations are not easily observable for smaller datasets but may become problematic when tested on production databases.
In addition, different query errors may have different performance implications. 
A missed column or join operation  may end up pulling too much data increasing the load on the database engine significantly. 
Some large models (such as LLM-based models) may also have considerable delay for inference, adding significant overhead to query execution. 
\section{BIS Benchmark Question design}
\label{section:questions}

We analyzed common business intelligence questions in our organization by checking the historical patterns and categorized the questions based on their intent and complexity of their relations with the databases. 
The nine most common  query categories identified are defined as follows:

\subsection*{Filter Queries}  
The user can choose to constrain the set of data in the analysis, usually based on a specific value they have chosen. Such filters might also optionally contain temporal constraints.
These type of queries have a basic filter structure and the key challenge for these queries is inferring the filter conditions accurately. 
The queries might also have implicit temporal constraints depending on the intent of the user.
The other points of complexity is the data type of the filters and usage of multiple filters combined with boolean operators. Some data types might be hard to infer. For example in some countries, postal code is an integer while in some others postal codes are strings.

\begin{description}
     \item \textbf{Example 1:} What are the sales in Dublin?
     \item \textbf{Example 2:} What are the sales in Dublin, London and Paris?
     \item \textbf{Template: }\mintsql{SELECT <columns> FROM <table>} \sqlbreak \mintsql{WHERE <conditions> AND <time_constraints>}
\end{description}

\subsection*{Aggregation and Group by Queries} 
The user can get a calculated summary of the data they have asked for, such as a sum, average, and count, which can be grouped by another value. These type of queries power the key indicator panels in \BI dashboards.
The main complexity of these types of queries is inference of aggregation function and group by columns.
Aggregation functions may be custom functions in different \BI scenarios. 
Therefore mapping to these custom aggregation functions might be challenging.
In addition, if the group by columns are inferred incorrectly, the query might not execute correctly.
Lastly, aggregations may also be nested in certain cases and the ordering of the aggregation functions in the nested representation may change the result.

\begin{description} 
    \item \textbf{Example 1:} Show me the average age of the ad user per city
    \item \textbf{Example 2:} Get the average age of the ad user per country and city
    \item \textbf{Template: }\mintsql{SELECT aggregator( <column> ) FROM <table>} \sqlbreak \mintsql{WHERE <conditions> AND  <time_constraints> GROUP BY <group_cols>}
\end{description}

\subsection*{Top/Bottom Selection Queries} The user can see the top or bottom X number of values for a metric, or rank a value in a metric. These type of queries can be associated with basic reports in \BI dashboards. The associated query of these questions can be complex and inference of multiple variables may create accuracy issues. 
Users might also want to see rankings change based on different filter selections. The system needs to handle this interactivity.

\begin{description}
    \item \textbf{Example 1: }What were the top 10 selling brands last year?
    \item \textbf{Example 2: }What is the lowest rated products?
    \item \textbf{Template:} \mintsql{SELECT aggregator( <column> ), <column> FROM <table>}\sqlbreak  \mintsql{WHERE <conditions> AND <time_constraints>}\sqlbreak \mintsql{ORDER BY aggregator( <column> ) LIMIT <X>}
\end{description}

\subsection*{Time period Queries} 

The user can ask for metrics from the most recent time period. These queries can generate either a single aggregation of the metric or a metric grouped based on intervals.
There are many formal or informal ways of specifying time constraints in natural language ranging from formal (isodates, Unix epoch) to very informal (recently, soon).
Moreover, the time constraint may indicate a range of time thresholds or a single upper/lower threshold.
Temporal constraints may also be explicitly or implicitly stated using natural language by the user.
For example when a user asks about the sales, the intent might include sales whole data, last month or last year.
NL2SQL should infer such hidden intents of the user to provide useful query output.
For these reasons, time constraint inference is quite challenging for NL2SQL models and these types of queries are not provided by current benchmarks to the best of our knowledge.

\begin{description}
    \item \textbf{Example 1:} Sales in last 2 weeks.
    \item \textbf{Example 2:} (\textit{formal temporal constraint}) What are the sales in Dublin last month?
    \item \textbf{Example 3:} (\textit{informal temporal constraint}) What are the sales in Dublin on 2024-07-01?
    \item \textbf{Template:} \mintsql{SELECT aggregator( <column> ), <column> FROM <table>}  \sqlbreak \mintsql{WHERE <conditions> AND <time_constraints>}
\end{description}

\subsection*{Comparison Queries} 

These queries are used to compare a metric between two groups. Usually the comparison is done between two entities and the resulting query is a combination of two queries. A \textbf{with} SQL expression can be used to generate two inner queries and an outer query can join these two findings. A  time constraint can also be added. Some models may do the operation in two stages initially extracting the data to compared in the first stage and merging the two queries in the second stage.
\begin{description} 
    \item \textbf{Example: }Compare sales of chocolate versus ice cream in 2022.
    \item \textbf{Template: }\mintsql{WITH (query 1) as t1, WITH (query 2) as t2}\sqlbreak \mintsql{SELECT <columns> FROM t1, t2 WHERE <joins>}
\end{description}

\subsection*{Trend Queries} 

These queries check the trend of a KPI or metric for a specified time period in set periods such as hours, days and weeks. Note that in real queries these limits or periods might be incomplete in certain cases and the model might have to infer these values intuitively. In addition converting timestamps to time intervals can be different for different databases. 
Trend is usually generated to build some histogram chart representation.
In such visual representations, time chunk aggregations and smoothing might also be challenging for the NL2SQL models in business applications since the parameters used for this functionality might change the trend function.

\begin{description}
    \item \textbf{Example:} Show me weekly revenue for Dublin in the last 3 months.
    \item \textbf{Template:} \mintsql{SELECT time_aggregator( <column> ), aggregator ( <column> )}\sqlbreak \mintsql{FROM <table> WHERE <time_constraints>}\sqlbreak \mintsql{ORDER BY  time_aggregator (<column>)}
\end{description}

\subsection*{Trend Comparison Queries}  

This category of questions tests the model's ability to generate queries with trend comparison across two time periods. Similar to comparison queries, the easiest way to query this category is usually by using the \mintsql{WITH} clause. Trend comparisons are essential in various contexts, such as business analytics, where understanding changes over time can inform strategic decisions. The \mintsql{WITH} clause helps in structuring these queries by allowing the creation of temporary result sets, which can then be joined or compared in the final query.

\begin{description}
    \item \textbf{Example:} Compare weekly revenue between this month and last month.
    \item \textbf{Template:} \mintsql{WITH (query 1) as t1, WITH (query 2) as t2}\sqlbreak \mintsql{SELECT <columns> FROM t1, t2 WHERE <joins>}
\end{description}

\subsection*{Multiple Tables Queries} 

This category of questions can only be answered after joining multiple tables. In SQL, there are different types of join operations. A model can infer the type of join operation based on domain knowledge or fallback to a default join type such as "outer join", to avoid missing data. Join operations can get especially tricky if multiple tables contain similar data such as different tables for predicted and actual metric records. In addition, join operation is risky since an incorrect join might pull cross product of rows from multiple tables.

\begin{description}
     \item \textbf{Example 1:} The revenue of the city with highest population in Germany. 
     \item \textbf{Sample  template:} \mintsql{SELECT <columns> FROM <tables> WHERE <joins>}
\end{description}

\subsection*{Percentage Queries} 

This category of questions are related to the percentage of some key metric in business. These query outputs are frequently used to generate charts to generate summaries for the key business metrics.
Models may either do the percentage calculation of a metric with a single SQL or evaluate the percentage of the metric in two stages.
\begin{description}
    \item \textbf{Example:} Sale shares per category of products.
    \item \textbf{Sample  template:} \mintsql{SELECT percentage(<column>) FROM <table> GROUP BY <columns>}
\end{description}

\section{BIS Benchmark Setup}
\label{section:benchsetup}
We designed our benchmark to overcome the challenges encountered by currently available NL2SQL benchmarks and support common \BI  question types through analyzing common questions of our organization's \BI application users.
We used a database schema frequently used for business intelligence scenarios to provide a more realistic benchmark with time based observations, value mapping and redundant definition of data (such as predicted and actual metric log values).
As discussed earlier, lack of temporal test data was a major limitation of earlier benchmarks making them insufficient to test many common \BI questions.
The sample database also contains some technical terms and abbreviations with associated mapping table. Questions in a non-Latin language highlight the challenges related to mixed use of the language within the context. Finally, we developed two evaluation metrics to overcome evaluation issues of the currently available metrics (described in detail in Section \ref{section:metric}).

\begin{table}[t] \footnotesize
\caption{Benchmark Question Count}
\label{table:benchmark-categories}
\begin{tabularx}{1\textwidth}{|X|X|r|} 
 \hline
 \textbf{Category} & \textbf{Explanation} & \textbf{Question Count} \\
 \hline
Filtering & Basic table filters &  30 \\
 \hline
Time period & NL time period checks & 40 \\
 \hline
Comparison & Comparison of two entities &  20\\
 \hline
Trend comparison & Comparison of 2 time period trends & 39 \\
 \hline
Multi table & Complex multi table queries &  36\\
 \hline
Rank & Ranking of variable of interest & 20 \\
 \hline
Percentage & Basic table filters &  26\\
 \hline
Aggregation & Aggregation of value &  14\\
 \hline
Language & Queries that search language specific constructs & 14 \\
 \hline
\end{tabularx}
\end{table} 

The proposed benchmark contains two databases. The first database contains real and predicted metrics of advertisement campaigns in 5 tables. The second database contains system operational data in 3 tables.  We define the questions in 9 categories as described  in Table \ref{table:benchmark-categories}. We provide sample SQLite databases for the tables with  sample business data. This set of tables are accessed multiple ways by the business analysts and provide a test bed for a set of business operations challenges in NL2SQL as shown in Section \ref{section:challenge}. The benchmark contains a mock database to help with evaluation of results for a sample NL2SQL model as well as true SQL results as curated by the analysts. In total, there are 239 questions to test an NL2SQL model in \BI scenarios. The majority of these categories with the exception of filtering and percentage could not be found in other open NL2SQL benchmarks. The benchmark comprising of question and SQL pairs, usage instructions, evaluation scripts and sample database can be downloaded from the the following Github repository: \url{https://github.com/boracaglayan/bis-nl2sql}.
A Python implementation of the similarity metrics explained in Section \ref{section:metric} is also provided.
Evaluation scripts require Python 3.9+ to run and more detailed instructions and required Python libraries are in the \mintsql{readme.md} and \mintsql{requirements.txt} file in the benchmark archive file respectively.

A complication related to time-associated questions, such as "yesterday" or "next 3 days," on a sample dataset is that the data is often generated for a short time period. Executing queries afterwards may yield no results if there is no data for that period, or conflicting results due to the change in timestamp anchor points, creating non-reproducible results. The sample databases contain time-tagged observations ranging from \mintsql{2023-01-02T00:00:00} to \mintsql{2023-01-17T00:00:00}. To reproduce results accurately, the current time of the system should be set to \mintsql{2023-01-17T00:00:00}.

A sample benchmark question is given as follows:

\begin{lstlisting}[language=json]
{
    "db_id": "benchmark_1",
    "query": "SELECT count(*) 
              FROM pre_ranking_filter_log 
                WHERE task=342111 
                  AND filter_key = 'o_rta_filter'",
    "question": "rta filtering count for task 342111?",
    "language": "en",
    "case_type": "filtering"
}
\end{lstlisting}

\begin{figure}[t]
    \centering
    \includegraphics[width=1\linewidth]{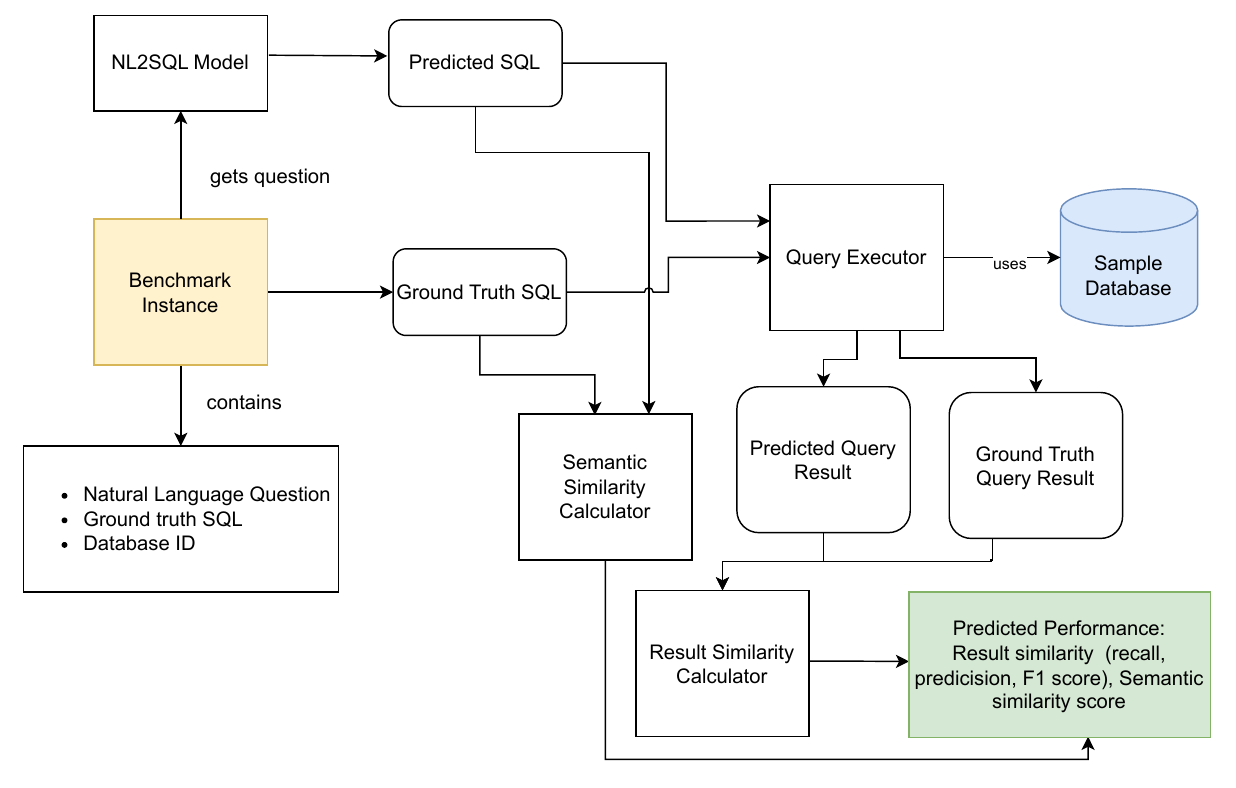}
    \caption{Calculation of result and semantic similarity for a benchmark instance.}
    \label{fig:pipeline_of_evaluation}
\end{figure}

In the sample benchmark question, db\_id key is used to specify the associated database schema. 
Query is the ground truth query generated by \BI system in production.
Question and language are the natural language question and language respectively and lastly the case type is the type of the question.
An SQLite database with sample data is created during evaluation and semantic similarity and result similarity metrics are calculated as explained in  Figure \ref{fig:pipeline_of_evaluation}. 
The benchmark is  NL2SQL model agnostic so any model, such as models  based on LLM APIs or other techniques, can be plugged easily.
Predicted SQL is generated by an NL2SQL model and the semantic similarity estimator estimates the semantic similarity between the two queries as explained in detail in Section \ref{sumsection:semmetric}.
If the predicted SQL is invalid semantic similarity is 0 by default since AST can only be generated from a valid SQL string. 
The SQL results are generated by executing the predicted and ground truth SQL queries on the associated benchmark database.
The SQL results are two data frames that can be compared to calculate the resulting similarity scores as explained in Section  \ref{sumsection:resmetric}.

\section{Evaluation Metric Design}
\label{section:metric}
Following a review of the issues of existing  performance measures of NL2SQL benchmarks as  discussed in Section \ref{section:challenge}, we defined two evaluation metrics namely SQL statement semantic similarity and SQL result partial similarity to assess the partial and structural similarity between predicted and ground truth SQL queries more effectively.

\subsection{SQL Statement Semantic Similarity}
\label{sumsection:semmetric}
\begin{figure}
    \centering
    \includegraphics[width=0.7\linewidth]{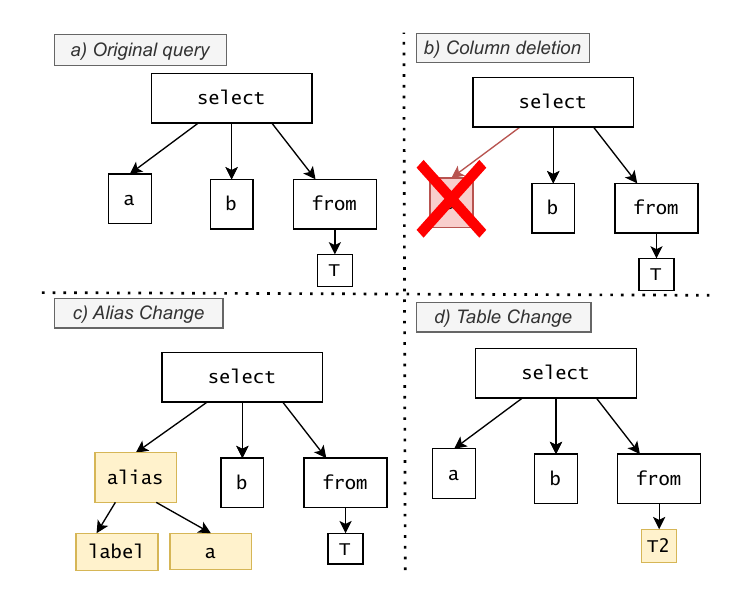}
    \caption{Sample Changes to a Test Query. The transformation type changes the query semantic similarity with different weights.}
    \label{fig:semsim-examples}
\end{figure}

SQL statement semantic similarity is computed by comparing the AST (abstract syntax tree) of two queries after transpilation to the same target database using Sqlglot\footnote{SQLglot: \url{https://github.com/tobymao/sqlglot}}. Transpilation is done before comparison to make sure similar SQL idioms and functions are compared with the same SQL dialect. Although SQL has various ISO and ANSI standards \cite{michels2018new}, some key functions and data structure representations are different between databases such as Clickhouse\footnote{Clickhouse: \url{https://clickhouse.com/}} and PostgreSQL\footnote{Postgresql: \url{https://www.postgresql.org/}}. 

Semantic similarity can be detected between queries with different ordering of select, filter, group by, order query components and gives a better approximation of model accuracy where testing execution accuracy is not practical. We can illustrate the semantic similarity calculation with a simple example: In Figure \ref{fig:semsim-examples}, we see 3 transformations for a simple query \mintsql{SELECT a,b FROM t}. The first transformation removes column \mintsql{a} and transforms the query to \mintsql{SELECT b FROM t}. This transformation would create a new SQL format but also has some structural similarities with the original query such as the co-occurrence of one column and accessing the same table. The second transformation adds a label to one of the columns in the original query a label to generate \mintsql{SELECT a as label, b FROM t}. This transformation generates the same output but only the column label would be different. Finally, the last transformation changes the query to access a different table as \mintsql{SELECT a, b FROM t2}. This transformation changes the accessed table. The change of this one element might cause a dramatic change in the SQL result and may reduce the semantic similarity of the queries dramatically. In our semantic similarity estimation algorithm, we treat these three changes with different weights.

In Algorithm \ref{alg:semsim} the calculation steps of semantic similarity is shown at a high level. 
Note that the AST generation and diff calculation is done using Sqlglot library. 
Semantic similarity  has a range of $ 0.0 \leq \text{semantic similarity} \leq 1.0 $.
The summary of the similarity calculation is as follows: 1) Keep and move operations for identical constructs are not penalized, 2) if the accessed table is changed similarity is automatically reduced to 0.0, 3) other changes in AST are counted. The total count is divided by all items in the AST diff list for normalization unless table access constraint is violated. 

\begin{figure}
    \centering
    \includegraphics[width=0.5\linewidth]{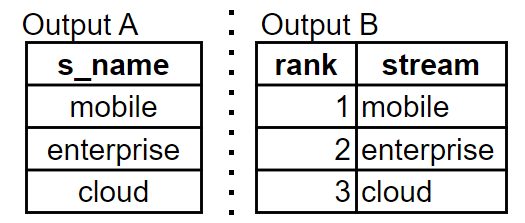}
    \caption{Two results for the query "what is the top 3 revenue streams?". The output A shows the ranking of revenue streams implicitly while the output B shows the ranking of revenue streams  explicitly. Semantic similarity performance measure does not over-penalize such output differences.}
    \label{fig:result-examples}
\end{figure}

\begin{figure}[t]
    \begin{algorithm}[H]
        \SetKwInOut{Input}{Input}
        \SetKwInOut{Output}{Output}
        \Input{query\_true, query\_predicted}
        \Output{semantic similarity s where $0<=s<=1$}
        \tcp{Transform the queries to AST and take their AST diff}
        ast\_diff = diff(parse(query\_true), parse(query\_predicted))\;
        size\_union = length(ast\_diff)\;
        diff\_count = 0\;
        \tcp{Diff is calculated differently based on change type. Table change diff is critical, move and keep type changes are trivial.}
        \For{element in ast\_diff}{
            \uIf {type(element) in (keep, move)}
            {continue}
            \uIf{type(element) in (delete, insert)}
            { \uIf {element.token == table}{size\_diff = size\_union\; break \;}
             \uElseIf {element.token == alias}{continue \;}
             \uElse {diff\_count += 1 \;}}
        }
        semantic\_similarity = min(size\_union, diff\_count) / size\_union\;
        return semantic\_similarity
        \caption{Semantic Similarity Estimation Algorithm}
        \label{alg:semsim}
    \end{algorithm}
\end{figure}

\subsection{SQL Result Partial Similarity}
\label{sumsection:resmetric}

SQL result similarity is based on the predicted and ground truth query result on a sample database.
For example if the query is the "What is top 3 revenue streams?", comparison of two results having ranks 1, 2 and 3 in one column explicitly or not would not change  the information content of the result completely for the user. Two outputs for such query can be seen in Figure \ref{fig:result-examples}.
In addition, one SQL result may re-label the columns and the other SQL result may keep the table column names. 

Columns of predicted and actual query results may have different labels and as long as the two column values exactly match we estimate there is a column match. No partial column match is possible since partial comparison of the value tuples of two result tuples would be problematic so there is either a perfect match between two columns or no match. If there are multiple columns in both predicted and actual query results, all the combinations are checked exhaustively to generate the highest possible similarity score. If one query produces $M$ columns and the other produces $N$ columns, we perform $M \times N$ comparisons by examining all possible pairs. This approach would cause issues if large results are compared with many columns and rows (>1K rows >100 columns) but for our sample database both column count and row is low for the \BI queries.
On a Intel 11370H laptop, all the result similarity scores can be computed under 1.5 minutes for evaluation of a model on the whole benchmark (< 1 second per comparison and not taking into account the model inference time).

Finally, column matches are aggregated to generate precision, recall and F1 score per instance.  The measures for SQL result similarity can be defined as follows per instance: Let $\mathbf{A}$ be the set of columns of the result for the predicted query, and $\mathbf{B}$ the set of columns of the result for ground truth query. Precision is given by; 

\begin{align}
P = Precision_{result} &= | A \cap B | / | A | 
\end{align}

Recall for an instance is given by, 

\begin{align}
R = Recall_{result} &= | (A \cup B  \setminus  A) | / |B| 
\end{align}

The harmonic mean of precision and recall gives the  F1 score: 

\begin{align}
F1_{result} &= 2/(1/P+1/R)\\
& = 2/(1 / (| A \cap B | / | A |) + 1 / (| (A \cup B  \setminus  A) | / |B|))
\end{align}

Finally, these measures are aggregated by taking the arithmetic mean across all the instances to generate a summary score. Aggregation can also be done per query category to debug or compare models per different \BI question categories.  These measures find partial similarity between predicted and true SQL query results with partially overlapping columns in the results instead of an exact match check strategy used by existing benchmarks.

\section{Ethics and Data Protection}

The data sources were carefully reviewed and structured to exclude any form of sensitive information, such as personal data. Furthermore, for security purposes, all company business-related terminology and values in the provided sample SQLite database were obfuscated. 

\section{Limitations}

SQL result similarity measures have potential problems for complex cases:

The first potential weakness is related to the test database. 
The calculation of result similarity requires executing the predicted and actual query on a test database, and any problems with data in the test database could create issues. 
For example, if there were no sales yesterday, aggregations such as \mintsql{SELECT count(price) FROM sales} and \mintsql{SELECT min(price) FROM sales} would have the same results. 
The test database should be verified to avoid such issues.

The second potential problem is related to column matching. Since we are checking exact match per column, if one column matches partially, the match for that column would be $0$.
This might penalize models with incorrect filter clauses too much, since just changing one \mintsql{WHERE} clause would reduce the result similarity to $0$.

The third potential weakness is related to computational complexity. If the test database is too large, calculating the result accuracy might require significant resources. The test database should be as small as possible to avoid this weakness.

To test the multilingual performance, we provide natural language questions in both Chinese and English as these are the most common languages in our organization. Most of the questions have both English and Chinese versions to highlight  language specific issues in models. The benchmark currently does not cover other languages. One of the critical aspects of \BI queries is the time constraints. We incorporated the time constraints for most of the questions but mapping the natural language representations of time constraints exhaustively in natural language with SQL datetime was not done within the scope of the work. 
We hope to overcome these weaknesses in the next version of our benchmark.

\section{Conclusion}

In this paper, we presented a novel \BIN-focused benchmark and two evaluation metrics for the estimation of NL2SQL performance in realistic scenarios.
We went through the challenges of the existing benchmarks for \BI domain and attempted to address these challenges by building a \BIN-specific benchmark and two novel evaluation metrics.
We believe business specific benchmarks provide a good complement to the large scale generic open benchmarks currently available to evaluate models in a given domain. 
As a future work, we plan to extend the number of questions in each category and the number of business domains in the next version of BIS to increase the coverage of the benchmark.

\bibliographystyle{splncs04}
\bibliography{bib}
\end{document}